\documentclass{article}
\usepackage{ijcai19}

\usepackage{times}
\usepackage{soul}
\usepackage{url}
\usepackage[hidelinks]{hyperref}
\usepackage[utf8]{inputenc}
\usepackage[small]{caption}
\usepackage{latexsym}
\usepackage{graphicx}
\usepackage{amsmath}
\usepackage{booktabs}
\usepackage{algorithm}
\usepackage{algorithmic}

\usepackage{subfigure}
\usepackage{helvet}  
\usepackage{courier}  
\usepackage{amssymb}
\usepackage{bm}
\usepackage{color}
\usepackage{enumitem}

\newcommand{\newcite}[1]{\citeauthor{#1} (\citeyear{#1})}

\title{Mimicking Human Process: Text Representation via Latent Semantic Clustering for Classification\footnote{This work was presented at 2nd Workshop on Humanizing AI (HAI) at IJCAI'19 in Macao, China}}

\author{
Xiaoye Tan$^1$
\and
Rui Yan$^{1,2}$\and
Chongyang Tao$^{2}$\And
Mingrui Wu$^3$
\affiliations
$^1$Center for Data Science, Peking University, Beijing, China\\
$^2$Institute of Computer Science and Technology, Peking University, Beijing, China\\
$^3$Alibaba Group, Seattle
\emails
\{txye, ruiyan, chongyangtao\}@pku.edu.cn,
wu.mingrui@yahoo.com
}

\hypersetup{draft}
\begin{document}

\maketitle

\begin{abstract}
Considering that words with different characteristic in the text have different importance for classification, grouping them together separately can strengthen the semantic expression of each part. Thus we propose a new text representation scheme by clustering words according to their latent semantics and composing them together to get a set of cluster vectors, which are then concatenated as the final text representation. Evaluation on five classification benchmarks proves the effectiveness of our method. We further conduct visualization analysis showing statistical clustering results and verifying the validity of our motivation.
\end{abstract}

\section{Introduction}
Text classification is an important task in natural language processing with many applications. Suitable text encoding scheme will benefit it a lot. Early studies use discrete and context-insensitive approaches such as TF-IDF~\cite{sparck1972statistical} and weighted  bag-of-words~\cite{joachims1998text}. Along with the prosperity of research on the distributed representation of words~\cite{mikolov2013efficient,bengio2003neural}, linear combination of word embeddings in a sentence~\cite{shen2018baseline} is widely used for classification. However, this method captures the information from individual words only. Many studies that make use of contextual information commonly adopt neural networks as an encoder to capture the semantic information of the text, including recurrent neural networks (RNN) \cite{sutskever2014sequence}, tree-structure recursive networks~\cite{Tai2015treelstm} and convolutional neural networks (CNN)~\cite{Kim2014ConvolutionalNN,kalchbrenner2014convolutional,conneau-EtAl:2017:EACLlong}. 

However, the aforementioned methods lack consideration on the semantic segmentation of text. It is intuitive that the words in text play different roles thus contribute to different groups of semantic functionality. Words like ``the'', ``an'', ``of'' act as connection between the syntax structures of the text and contribute little for classification. Other words like ``internet'', ``linux'' or ``wireless'' indicating a specific domain (science) tend to have a huge impact on the classification performance. Some recent studies attempt to capture different types of word importance via hierarchical attention \cite{yang2016HAN} or multi-head self-attention \cite{lin2017structured}. However, these attention mechanisms still only consider the relative importance of separate words, which does not group words with similar semantics together to strengthen the semantic expression of the whole text. 

As classification is a word sensitive task, people tend to integrate all related words when making decisions about the category of a sentence. Therefore, in this paper, we propose to augment text representation from a higher level: cluster level. Concretely, we divide the words in a text into different latent semantic clusters and get cluster representations by combining contextual embeddings of the words together according to their cluster probability distribution. The cluster representations are then concatenated as the final representation of the text. We further introduce two regularization terms to better guide the clustering process. Considering that not all semantic clusters contain useful information for classification, we design a gating mechanism to dynamically control their contributions for classification.

Experiments are conducted on five standard benchmark datasets of text classification. Quantitative results show that our method outperforms or is at least on a par with the state-of-the-art methods. We also perform visualized and statistical analysis on the intermediate word clustering results which proves the effectiveness of the clustering process. 

In summary, the contributions of this paper include:
\begin{itemize}[nosep,leftmargin=1em,labelwidth=*,align=left]
\item We propose an intuitive architecture for text classification with a novel semantic clustering process for better capturing distant topical information in text representation. The semantic clusters in our framework are automatically calculated on-the-fly instead of being fitted in advance.

\item Due to the probabilistic nature of soft semantic clustering, we introduce two regularization schemes to better guide the behaviors of our model. 
\item We conduct extensive experiments to evaluate our proposed method on five text classification benchmarks. Results show that our model could obtain competitive performance compared with state-of-the-art approaches.
\item We provide statistical and visualization analysis on cluster distribution captured by our learned model further corroborating our motivation.
\end{itemize}
\begin{figure}
\centering
\includegraphics[height=0.9\linewidth]{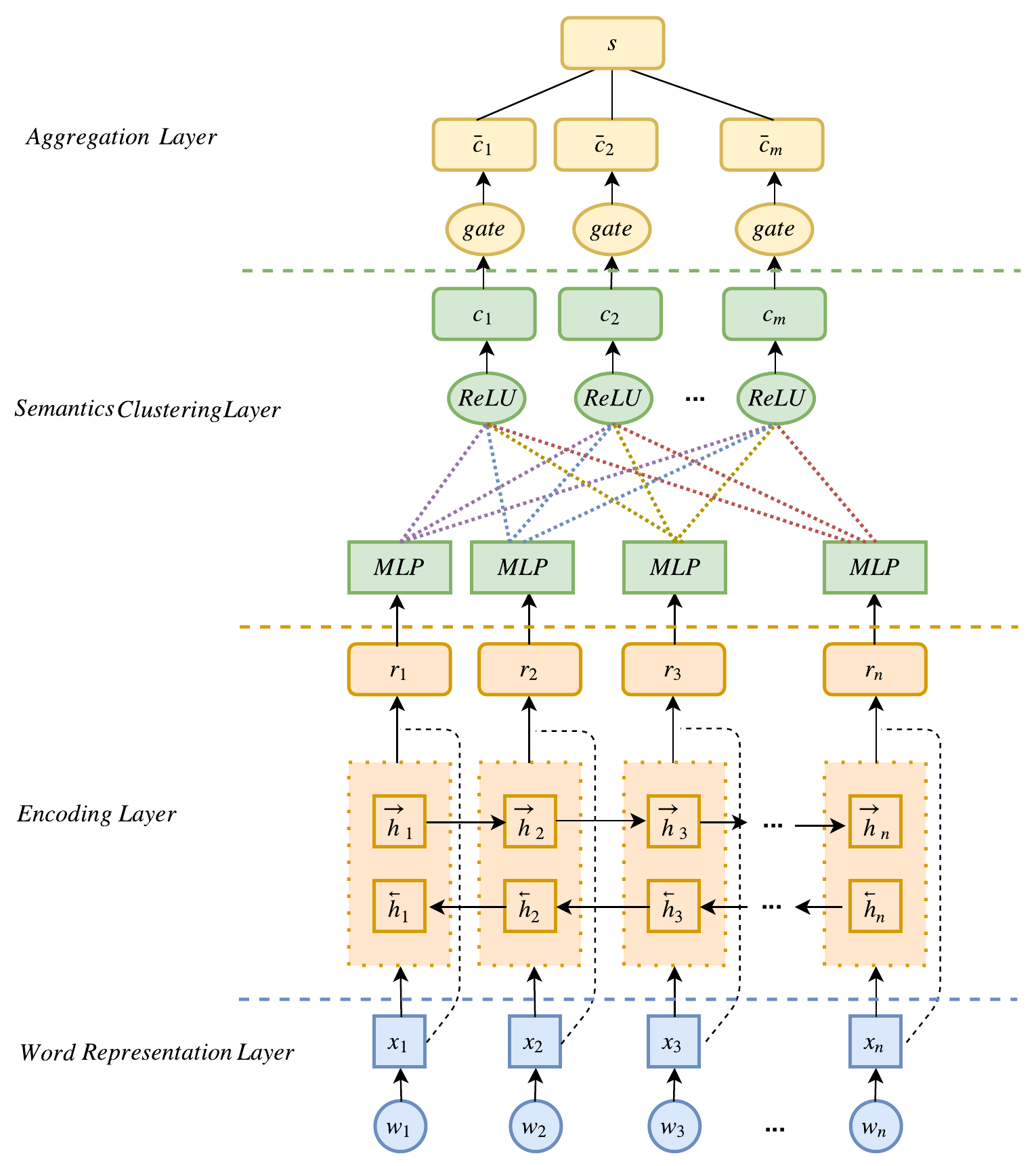}
\vspace{-2mm}
\caption{The framework of our model.}
\vspace{-4mm}
\label{fig:framework}
\end{figure}
\section{Model}
We propose a latent semantic clustering representation (LSCR) framework as shown in
Figure~\ref{fig:framework} consisting of four parts: 1) the word representation layer at the bottom converts words into vector representations (embeddings); 2) the encoding layer transforms a series of word embeddings to their corresponding hidden vector representations; 3) the semantics clustering layer attributes all the words into different clusters and composes them together respectively so as to get a set of cluster representations; 4) the aggregation layer combines those cluster vectors into a single vector as the final representation of the text.  

Formally, given a text consisting of $n$ words $(w_1, w_2, \cdots, w_n)$, the word representation layer converts them to their corresponding word embeddings, represented as $\bm X=(\bm x_1, \bm x_2, \cdots, \bm x_n)$. Then in encoding layer, we employ a bi-directional LSTM~\cite{schuster1997bidirectional} as the encoder to aggregate information along the word sequence getting final states $\bm h_t = [\overrightarrow{\bm h}_t;\overleftarrow{\bm h}_t]$. We concatenate the hidden state from the encoder with the initial word embedding to enrich the representations. The output of the $t$-th word from this layer takes the form: $\bm r_t = [\bm x_t;\bm h_t]$.

The final representation of words generated from this layer is defined as $\bm R= (\bm r_1, \bm r_2, \cdots, \bm r_n)$. In the semantics clustering layer, suppose that words from the text could be assigned to $m$ semantic clusters, where $m$ is a tunable hyperparameter. For each word, we employ a MLP to determine the probabilities of dividing it into every cluster, defined as: 
\begin{equation} \label{prob}
    \bm A= f_1(\bm W_2 \cdot f_2(\bm W_1 \cdot \bm R + \bm b_1) + \bm b_2)
\end{equation}
We use the softmax function for $f_1$, and ReLU for $f_2$.
Concretely, $A_{i, j}$ indicates the probability of the $j$-th word being clustered into the $i$-th cluster. For each word $w_j$ of the text, $\sum_{i=1}^m A_{i,j}=1$. After getting the probabilities, the vector representation of the $i$-th cluster is given by the weighted sum of the contextualized word representations ($ \bm R$) in the text, followed by a nonlinear transformation. The process is formulated as:

\begin{equation}
\bm C=\text{ReLU}(\bm W_s(\bm A \cdot \bm R) + \bm b_s)
\end{equation}
The $i$-th row of $\bm C$, $\bm c_i$, refers to the vector representation of the $i$-th cluster. Considering that not all clusters are helpful for text classification. There may exist redundant clusters that contain little or irrelevant information for the tasks. Therefore, in the aggregation layer, we add a gating mechanism on the cluster vector to control the information flow. Concretely, the gate takes a cluster vector $\bm c_i$ as input and output a gate vector $\bm g_i$:
\begin{equation} \label{gate}
\bm g_i = \sigma(\bm W_g\bm c_i + \bm b_g)
\end{equation}
We do the same operation on other cluster vectors as well, leading to a series of gate vectors $\bm G = (\bm g_1, \bm g_2, \cdots, \bm g_m)$ from the cluster vectors. Then the gated cluster vectors $\bar{\bm C} = ( \bar{\bm c}_1,  \bar{\bm c}_2, \cdots, \bar{\bm c}_m)$ are calculated as:
\begin{equation}
\bar{\bm C} = \bm G \odot \bm C
\end{equation}

At last, we concatenate vector representations of all the semantic clusters to form the text representation:$\bm s =[\bar{\bm c}_1, \bar{\bm c}_2, \cdots ,\bar{\bm c}_m]$.
For classification, the text representation $\bm s$ is followed by a simple classifier which includes a fully connected hidden layer and a softmax output layer to get the predicted class distribution $\bm y$. The basic loss function is the cross-entropy loss $\mathcal{L}$  between the ground truth distribution and the prediction.

\subsection{Regularization Terms}
Due to the probabilistic nature of our soft semantic clustering scheme, it is natural to integrate probabilistic prior knowledge to control and regularize the model learning process. We consider two regularization terms in word level and class level. In the semantics clustering layer, we get $\bm a_i = (a_{i1}, a_{i2}, \cdots, a_{im})$ indicating the probability distribution of the $i$-th word to each cluster. The \emph{word-level entropy regularization} term is defined as:
\vspace{-1.5mm}
\begin{equation} \label{word}
\mathcal{L}_{word} = -\sum_{t=1}^N\sum_{k=1}^{m}a_{tk}\log(a_{tk})
\vspace{-1mm}
\end{equation}
We expect the probability distribution for a specific word over all the clusters is sparse, which means a word would be attributed to only one or few clusters with greater probability instead of being evenly distributed to all clusters. Thus our optimization goal is to minimize the word-level entropy. 

Another class-level regularization term is specifically designed for text classification. Suppose there is a vector $\bm v_{c_i}$ indicating the $i$-th class' probability distribution over $m$ clusters, which is calculated by averaging the cluster probability distribution of the text belonging to $i$-th class within a mini-batch during training. We take the average of the word's cluster probability distribution $\bm a$ in the text as the text-level cluster probability distribution $\bm v_s = \frac{1}{N_w}\sum_{i=1}^{N_w} \bm a_i$, 
where $N_w$ is the number of words in the text.
Thus the $i$-th class' cluster probability distribution is:
\vspace{-1.5mm}
\begin{equation} \label{classprob}
\bm v_{c_i} = \frac{1}{N_{c_i}}\sum_{k=1}^{N_{c_i}} \bm v_{s_k}
\vspace{-1.5mm}
\end{equation}
where $N_{c_i}$ is the number of samples belonging to $i$-th class in a mini-batch, $\bm v_{s_k}$ indicates the $k$-th text-level cluster probability distribution. We hope that different cluster can capture different category related semantics. Thus the distribution between every two classes needs to be different. We take an intuitive and practical method that we expect the peak value of different class-level distribution exists in different cluster. To implement this, we add the maximum value of all the class-level distributions and expect the summation greater. The \emph{class-level regularization} term is defined as:
\vspace{-1.5mm}
\begin{equation}
\mathcal{L}_{class} = \sum_{i=1}^{m}\max_{j=1:N_C}(\bm v_{c_j}^i) 
\vspace{-1.5mm}
\end{equation} 
where $N_C$ is the number of category and $\bm v_{c_j}^i$ means the $i$-th dimension (corresponding to $i$-th cluster) in $j$-th class probability distribution vector. The larger the summation, the distributions between classes tend to be more different. 
The final objective function for classification is defined as:
\vspace{-1.5mm}
\begin{equation}
    \begin{aligned}
        \mathcal{L}_{total}=\frac{1}{N}\sum_{i=1}^N(\mathcal{L} +\lambda_1 \mathcal{L}_{word}) -\lambda_2 \mathcal{L}_{class}
   \end{aligned}
\vspace{-1.5mm}
\end{equation}
where $N$ is the number of samples in a mini-batch. The training objective is to minimize $\mathcal{L}_{total}$.
\begin{table}[!t] 
\centering
\resizebox{\linewidth}{!}{
    \begin{tabular}{c|c|c|c|c|c}
    \toprule[1pt]
    Dataset & \# Classes & \# Training & \# Testing & \# AvWords & \# MaxWords\\ 
    \hline
    AGNews  &4  &120K  &7.6K  &45   &208    \\ 
    Yah.A.  &10  &1.4M  &60K   &104   &146  \\ 
    Yelp P. &2  &560K  &38K   &153   &301    \\ 
    Yelp F. &5    &650K &50K &155   &501   \\ 
    DBPedia &14     &560K     &70K   &55   &151 \\ 
    \bottomrule[1pt]
    \end{tabular}
}
\caption{Data statistic on the five benchmarks.}
\label{tab:statistic}
\vspace{-4mm}
\end{table}

\begin{table*}[!htbp]
\centering
\resizebox{0.75\linewidth}{!}{
    \begin{tabular}{l|ccccc}
    \toprule[1pt]
    Method & AGNews & Yah.A. & DBPedia & Yelp P. & Yelp F.\\ 
    \hline
    n-gram TF-IDF \cite{zhang2015character}  &92.4  &68.5	&98.7	&95.4	&54.8    \\
    BoW \cite{zhang2015character}  &88.8	&68.9	&96.6	&92.2	&58.0     \\
    \hline
    FastText \cite{joulin2016bag} &92.5	&72.3	&98.6	&95.7	&63.9     \\ 
    SWEM \cite{shen2018baseline} &92.2	&73.5	&98.4	&93.8	&61.1     \\
    \hline
    DeepCNN(29 Layer)\cite{conneau-EtAl:2017:EACLlong}  &91.3 &73.4	&98.7	&95.7	&64.3    \\ 
    Small word CNN \cite{zhang2015character}    &89.1 &70.0 &98.2 &94.5 &58.6 \\ 
    Large word CNN \cite{zhang2015character}   &91.5 &70.9 &98.3 &95.1 &59.5 \\
    Dynamic-Pool$^\diamond$ \cite{kalchbrenner2014convolutional} &91.3	&-	&98.6	&95.7	&63.0 \\
    Densely Connected CNN \cite{wang2018densely} &\textbf{93.6}	&-	&\textbf{99.2}	&96.5	&66.0 \\ \hline
    LSTM \cite{zhang2015character} &86.1	&70.8	&98.6	&94.7	&58.2 \\
    char-CRNN \cite{xiao2016efficient} &91.4  &71.7 &98.6 &94.5 &61.8 \\
    ID-LSTM \cite{zhang2018learning} &92.2 &- &- &- &- \\
    HS-LSTM \cite{zhang2018learning} &92.5	&-	&-	&-	&- \\
    \hline
    Self-Attentive$^\diamond$ \cite{lin2017structured}  &91.5 &-	&98.3	&94.9	&63.4 \\
    \hline
    LEAM \cite{wang2018joint} &92.5	&77.4	&99.0	&95.3	&64.1 \\
    W.C.region.emb \cite{qiaoanew} &92.8 &73.7	&98.9	&96.4	&64.9 \\
    \hline
    LSCR (our work) & \textbf{93.6}	& \textbf{78.2}$^{*}$	& 99.1	&\textbf{96.6}$^{*}$	&\textbf{67.7}$^{*}$ \\    
    \bottomrule[1pt]
    \end{tabular}
}
\vspace{-3mm}
\caption{Accuracy of all the models on the five datasets. The result marked with $\diamond$ is re-printed from \protect \newcite{wang2018densely}.}
\label{tab:mainresult}
\vspace{-5mm}
\end{table*}

\section{Experiment}
\subsection{Datasets}
To evaluate the effectiveness of our proposed model, we conduct experiments on the text classification task. We test our model on five standard benchmark datasets (\textbf{AGNews}, \textbf{DBPedia}, \textbf{Yahoo! Answers}, \textbf{Yelp P.}, \textbf{Yelp F.}) including topic classification, sentiment classification and ontology classification as in~\cite{zhang2015character}. \textbf{AGNews} is a topic classification dataset that contains 4 categories: world, business, sports and science. \textbf{DBPedia} ontology dataset is constructed by choosing 14 non-overlapping classes from DBPedia 2014. The fields used for this dataset contain the title and abstract of each Wikipedia article. \textbf{Yahoo! Answers} is a 10-categories topic classification dataset obtained through the Yahoo! Webscope program. The fields include question title, question content and best answer. \textbf{Yelp P.} and \textbf{Yelp F.} are Yelp reviews obtained from the Yelp Dataset Challenge in 2015. Yelp p. predicts a polarity label by considering stars 1 and 2 as negative, 3 and 4 as positive. Yelp F. predicts the full number of stars from 1 to 5. We use the preprocessed datasets published by \cite{wang2018joint} and the summary statistics of the data are shown in Table \ref{tab:statistic}. 

\subsection{Compared Models}
We compare our models with different types of baseline models: traditional feature based models i.e. n-gram TF-IDF and bag-of-words (BoW)~\cite{zhang2015character}; word embedding based models such as FastText~\cite{joulin2016bag} and SWEM~\cite{shen2018baseline}; RNNs like LSTM~\cite{zhang2015character}; reinforcement learning based models including ID-LSTM and HS-LSTM~\cite{zhang2018learning}; CNNs consisting of DeepCNN~\cite{conneau-EtAl:2017:EACLlong}, small/large word CNN~\cite{zhang2015character}, CNN with dynamic pooling~\cite{kalchbrenner2014convolutional} and densely connected CNN with multi-scale feature attention~\cite{wang2018densely}; CNN combined with RNN~\cite{xiao2016efficient}; self-attentive based model~\cite{lin2017structured}; other models specifically designed for classification~\cite{wang2018joint,qiaoanew}.

\subsection{Implementation Details}
For word representation, we use the pre-trained 300-dimensional GloVe word embeddings~\cite{Pennington2014glove}. They are also updated with other parameters during training. We split 10\% samples from the training set as the validation set and tuned the hyper parameters on validation set. The input texts are padded to the maximum length appeared in the training set. In the encoding layer, the hidden state of the bi-LSTM is set to 300 dimensions for each direction. The MLP hidden units are 800 for semantic clustering. The dimension of the clustering vectors is set to 600. The cluster number is set to 8 on AGNews and 10 on other datasets. The MLP used for classification has 1000 hidden units. The coefficients of the regularization terms are set to 0.001. We adopt Adam as the optimizer with a learning rate of 0.0005. The batch size is 64. Our models are implemented with Tensorflow~\cite{abadi2016tensorflow} and trained on one NVIDIA 1080Ti GPU. For all datasets, training converges within 4 epochs. We will release our codes later.

\begin{figure*}[h!]
  \centering
    \subfigure[\small{A sentence about business.}] { \label{fig:business1}
        \includegraphics[height=23mm]{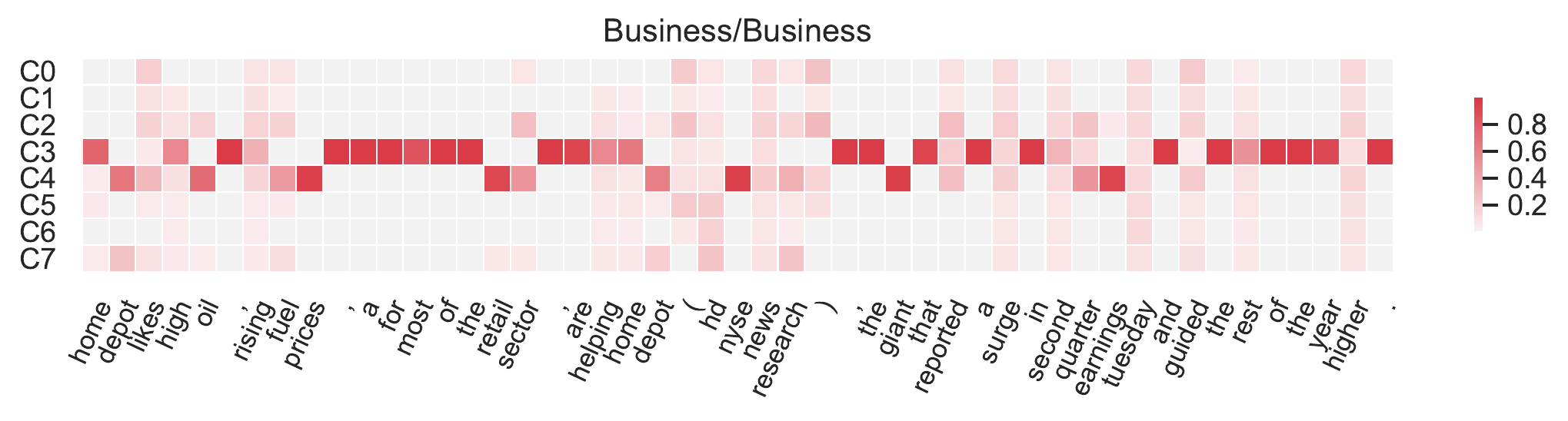}
    }  \hspace{-3mm}
    \subfigure[\small{A sentence about science.}] { \label{fig:science1}
        \includegraphics[height=23mm]{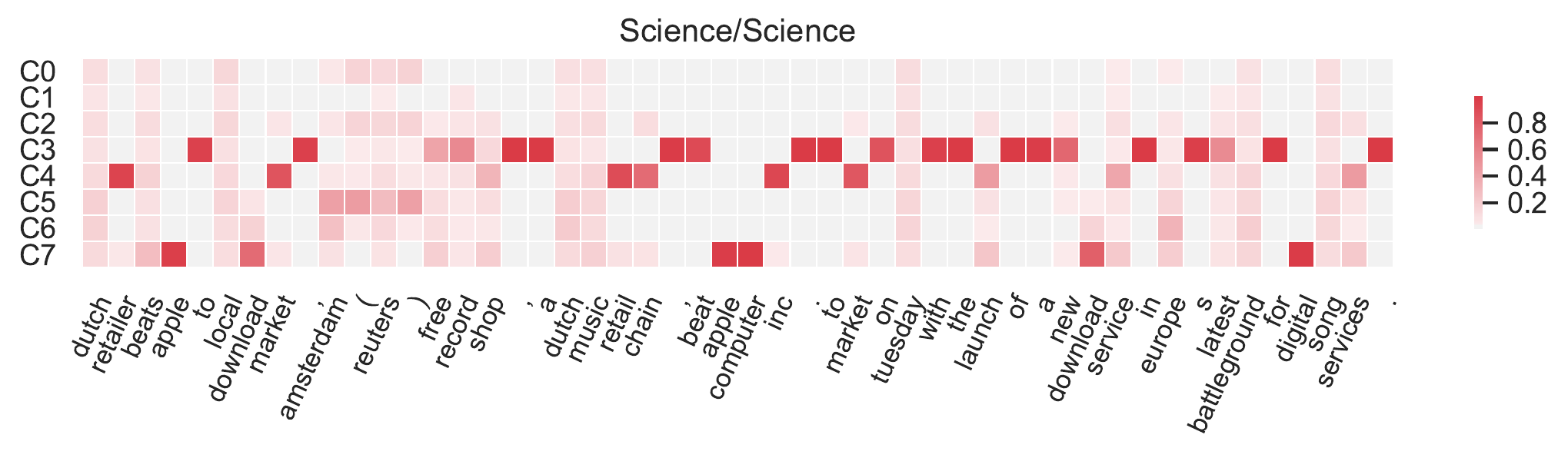}
    }
    \vspace{-3mm}
  \caption{The visualization on AGNews about topic classification. The heat map shows the distribution of words among different clusters in the text. The X-axis are words in text. The Y-axis represents the clusters. The title of each figure indicates the predicted class / the ground truth. } 
  \label{fig:topic_classify}
  \vspace{-5mm}
\end{figure*} 
\subsection{Evaluation Results}
We compared our method to several state-of-the-art baselines with respect to accuracy, the same evaluation protocol with \cite{zhang2015character} who released these datasets. We present the experimental results in Table \ref{tab:mainresult}.  The overall performance is competitive. It can be seen from the table that our method improves the best performance by 1.0\%, 0.1\% and 2.5\% on Yah A., Yelp P. and Yelp F respectively and is comparable on AGNews. Compared with the baselines, our results exceed the traditional feature-based models (n-gram TF-IDF, BoW), the word embedding based representation models (FastText, SWEM) and LEAM by a large margin. Furthermore, we gain great improvement over the LSTM-based models.
Though our model and the self-attentive model all aim at getting multiple vectors indicating different semantic information, as our model gather the words into clusters to enrich the representations, we achieve a better performance over theirs.  
Compared with the deep CNNs, our shallow model with a relatively simpler structure outperforms them as well. The models proposed by \cite{wang2018densely} and \cite{qiaoanew} aim at capturing features from different regions. Our results are on par with \cite{wang2018densely} and win over \cite{qiaoanew}.

\begin{figure}[t!]
  \centering
    \subfigure[\scriptsize{Cluster 3.}] { \label{fig:cluster3}
        \includegraphics[height=38mm]{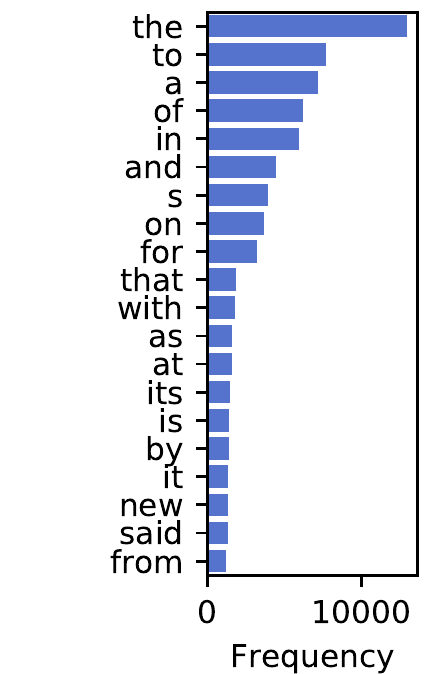}
    } \hspace{-3mm}
    \subfigure[\scriptsize{Cluster 4.}] {\label{fig:cluster4}
        \includegraphics[height=38mm]{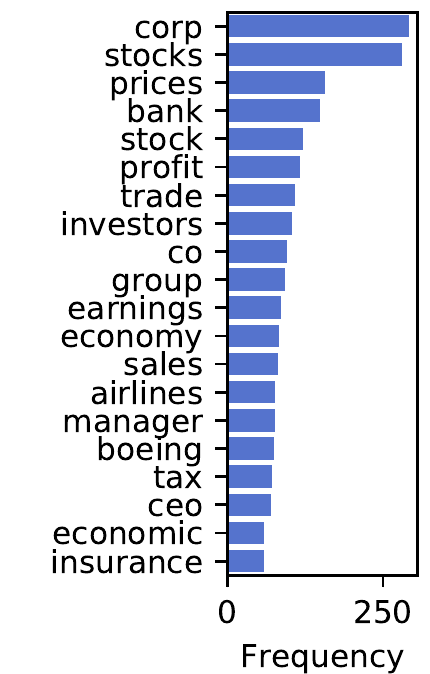}
    }  \hspace{-3mm}
    \subfigure[\scriptsize{Cluster 7.}] { \label{fig:cluster7}
        \includegraphics[height=38mm]{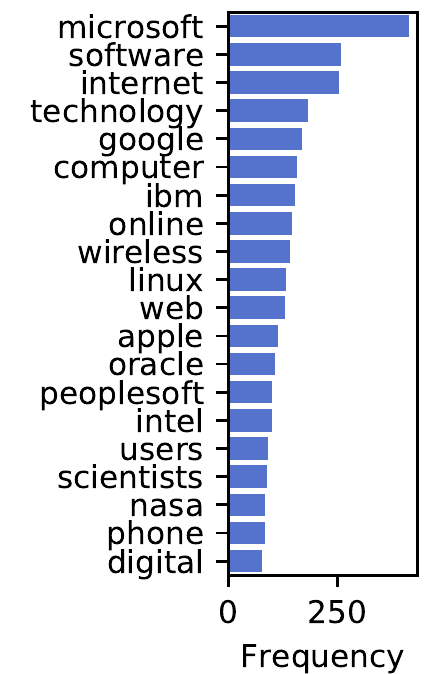}
    }  
    \vspace{-4mm}
  \caption{The statistical clustering results on AGNews in Cluster 3, 4 and 7.}
  \label{fig:top_20}
  \vspace{-4mm}
\end{figure} 
\subsection{Discussions}
\paragraph{Analysis on Clustering Results}
In this section, we take out the intermediate clustering results calculated when we test on AGNews which is a 4-category topic classification dataset including science, business, sports, and world. 

Firstly, we visualize the distribution of words in clusters using heat map as shown in Figure \ref{fig:topic_classify}. Each column means the probabilities of a word attributed to the clusters which sum up to 1. We can see that most of the meaningless words, high frequency words and punctuation are divided into cluster 3, like \textit{``a"}, \textit{``for"}, \textit{``that"}, etc. The text in Figure \ref{fig:business1} belongs to business category and the model predicts it right. We can find that the words about business are in cluster 4 such as \textit{``depot"}, \textit{``oil"}, \textit{``prices"}. Likewise, from Figure \ref{fig:science1} we find words about science are assigned to cluster 7 like \textit{``apple"}, \textit{``download"}, \textit{``computer"}. In Figure \ref{fig:science1}, the text contains words about science and business, which are divided into their corresponding clusters separately. The words like \textit{``beats"}, \textit{``second"}, \textit{``local"}, which are domain independent nouns or adjectives or verbs have a more average probability distribution among all clusters. 

We further make statistics about the clustering results over all the words. To be specific, we count the frequency of all the words being assigned to each cluster. A word is assigned to some cluster means that it is attributed to which with maximum probability. For each cluster, we sort the belonged words according to their frequency. We display the top 20 words in cluster 4, cluster 7 and cluster 3 as shown in Figure \ref{fig:top_20}. We can see that the words in cluster 4 are about business while the words in cluster 7 are science related and in cluster 3, the words are almost meaningless for classification. They are inconsistent with the cluster phenomenon in the heat map.

From the above visualization we have the following observations: first, words that indicate the same topic or with the similar semantics are divided into the same clusters; second, different categories correspond to different clusters; third, representative keywords have much greater probabilities (deeper color in heat map) in the specific cluster. These results exactly correspond to the motivation of the two regularization terms.

\begin{figure}[t!]
\includegraphics[width=0.9\linewidth]{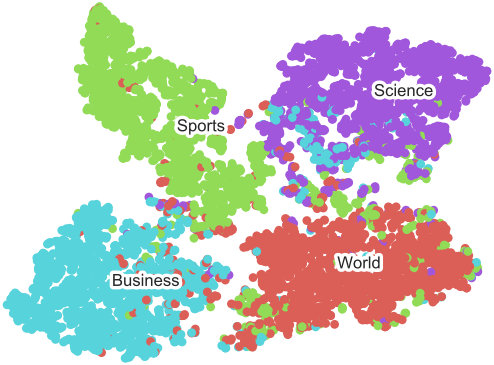}
\caption{t-SNE plot of the intermediate clustering probability distribution vector of text on AGNews test set.}
\label{fig:tsne}
\end{figure}

To evaluate the relevance between the clustering distribution and the classification results, we utilize t-SNE \cite{maaten2008visualizing} to visualize the text-level cluster probability distribution on a two-dimensional map as shown in Figure \ref{fig:tsne}. Each color represents a different class. The point clouds are the text-level cluster probability distribution calculated by averaging the words' cluster probability distribution\footnote{The cluster probability distribution is the intermediate result during testing. The initial dimension of the points is the cluster number. Each dimension represents the probability that the text belongs to a cluster.}. The location of the class label is the median of all the points belonging to the class. As can be seen, samples with the same class can be grouped together according to their text-level cluster probability distribution and the boundary between different class is obvious, which again demonstrates the strong relevance between the clustering distribution and the classification results.

\begin{figure}[!t]
  \centering
  \vspace{-3mm}
    \subfigure[\footnotesize{Performance across different cluster numbers.}] { \label{fig:cluster_trend}
        \includegraphics[height=0.35\linewidth]{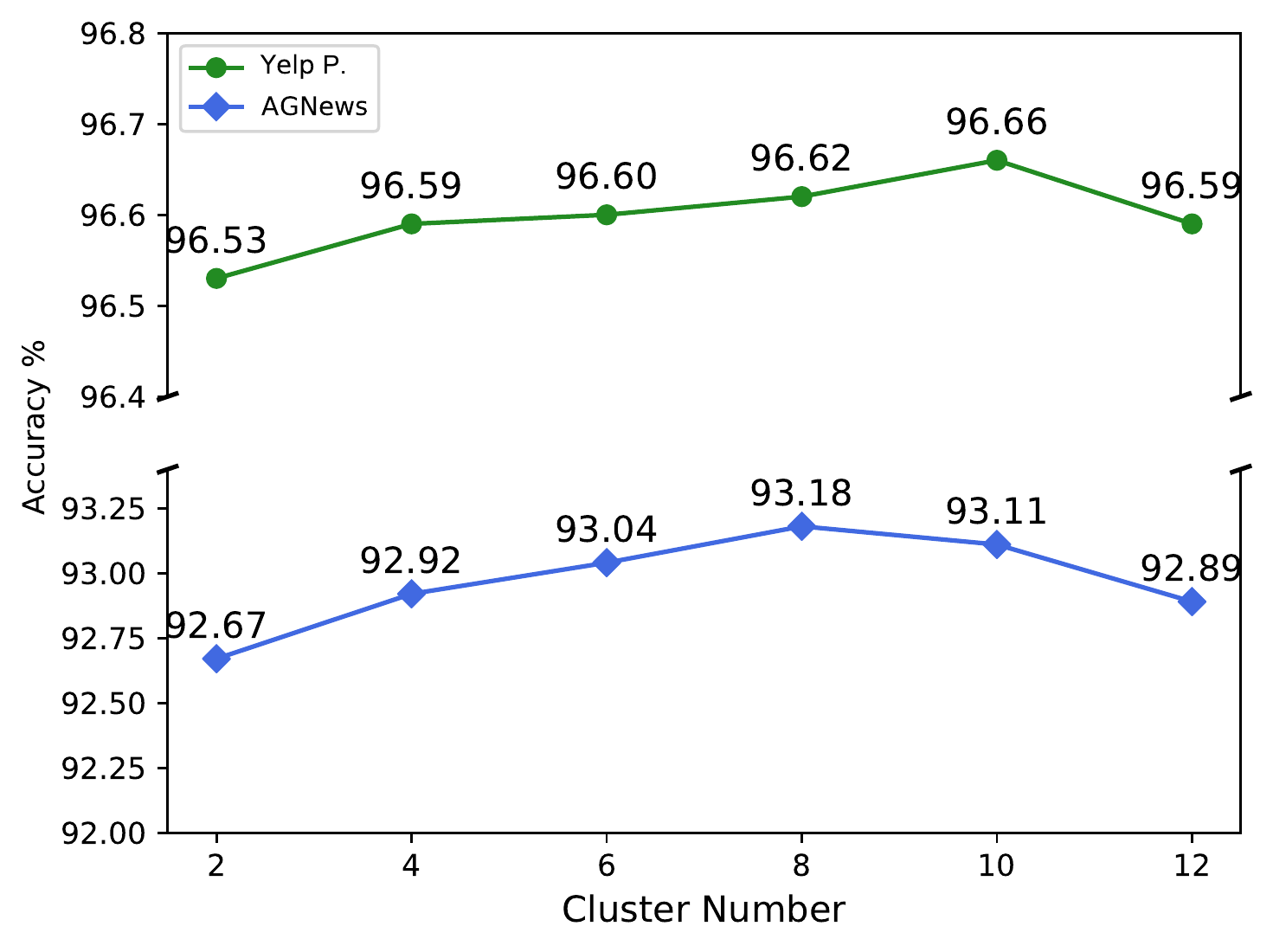}
    }  \hspace{-1mm}
    \subfigure[\footnotesize{Performance across different sentence length.}] { \label{fig:len_trend}
        \includegraphics[height=0.35\linewidth]{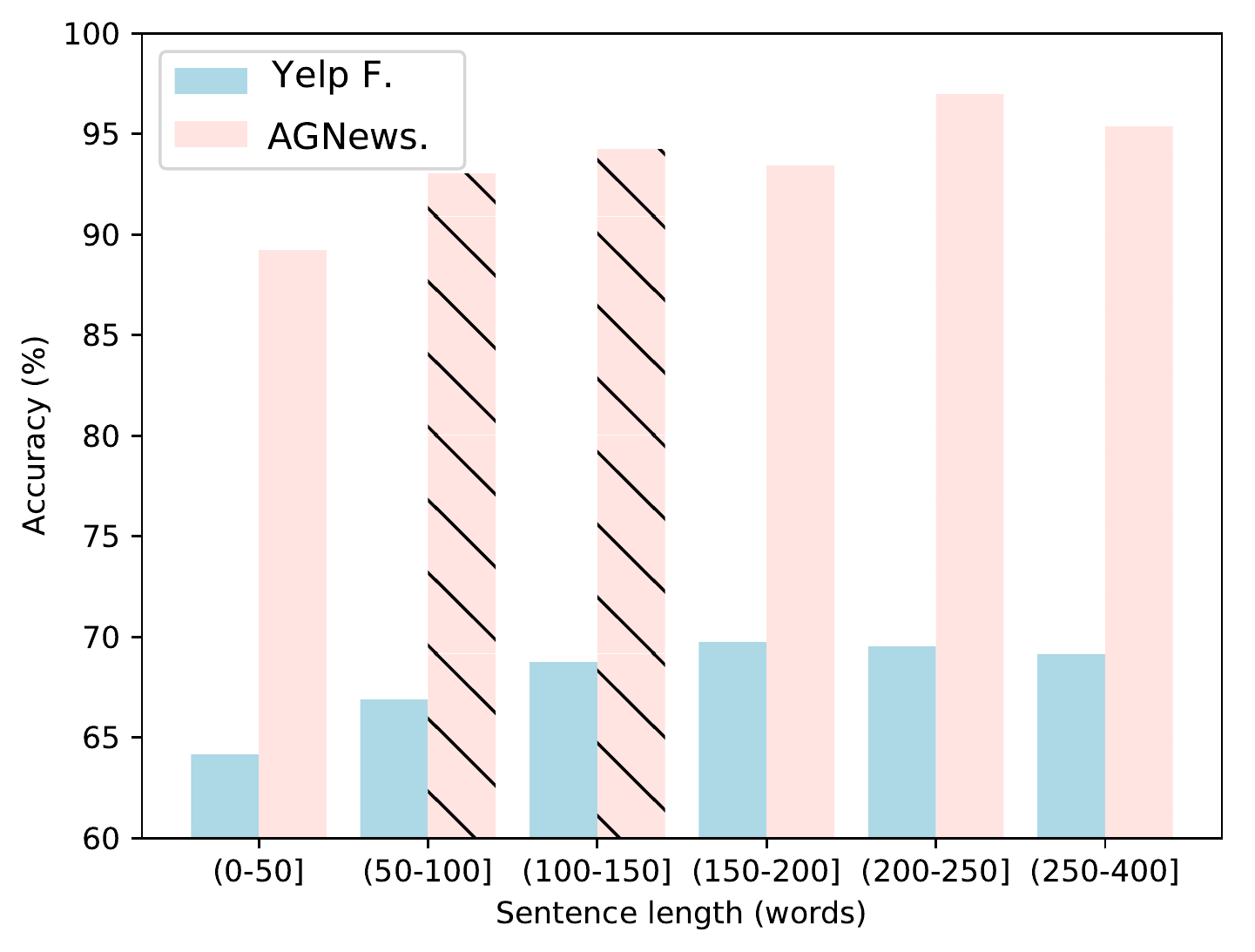}
    }
\vspace{-3mm}
  \caption{Quantitative analysis on cluster number and sentence length. }
  \vspace{-5mm}
  \label{fig:lenth}
\end{figure}

\paragraph{Analysis on The Number of Semantic Clusters.} 
As the number of semantic clusters is a tuned hyper parameter, we also analyze how performance is influenced by the number of semantic clusters by conducting experiments on AGNews and Yelp P., varying the cluster number $m$ among \{2,4,6,8,10,12\}. From Figure \ref{fig:cluster_trend} we can find that the accuracy increases as
the number of clusters increase and begins to drop after reaching the upper limit. Obviously, the cluster number does not in line with the class number to gain the best performance.

\paragraph{Analysis on Different Text Length.}
As the text length in datasets varies, we visualize how test accuracy changes with different text length. We perform experiments on AGNews and Yelp F. while the former has shorter text length than the latter. We divide the text length into 6 intervals according to the length scale. As Figure \ref{fig:len_trend} shows, our model performs better on relatively longer texts. With the increase in text length, our model tends to gather more information from the text. This visualization is in line with the overall performance of our model that it performs better on Yah.A., Yelp P. and Yelp F. rather than AGNews and DBPedia as the former three datasets have longer average text length. 

\vspace{-1.5mm}
\section{Related Work}
Text representation is an important and fundamental step in modern natural language processing, especially with the current development of approaches based on deep neural networks. bi-LSTM~\cite{schuster1997bidirectional} is a widely-used representation model which conveys information in both directions. Although it alleviates the problem of information vanishment due to the increase of sentence length, our framework can further integrate information by clustering words with similar semantics and give a visual explanation for the results. Another popular representation models are those based on CNN \cite{zhang2015character,kalchbrenner2014convolutional,conneau-EtAl:2017:EACLlong} which can capture the word features locally while our model can break the limitation of distance. A burgeoning and effective model is the structured self-attentive sentence embedding model proposed by \cite{lin2017structured}. They propose a multi-head self-attention mechanism mapping sentence into different semantic spaces and get sentence representation for each semantic space through attention summation. However, they focus on extracting different aspects of a sentence based on the automatically learned relative importance instead of composing similar aspects together to strengthen the information of each part as our work does. 
Except for the above representation models, there are other several models specifically designed for classification.  \newcite{wang2018joint} takes the label information into consideration by jointly embedding the word and label in the same latent space. \newcite{wang2018densely} uses a densely connected CNN to capture variable n-gram features and adopts multi-scale feature attention to adaptively select multi-scale features. \newcite{qiaoanew} utilizes the information of words’ relative positions and local context to produce region embeddings for classification.

\vspace{-1.5mm}
\section{Conclusion and Future Work}
In this paper, we propose to transform the flat word level text representation to a higher cluster level text representation for classification. We cluster words due to their semantics contained in their contextualized vectors gotten from the concatenation of the initial word embeddings and the outputs of the bi-LSTM encoder. We further introduce regularization schemes over words and classes to guide the clustering process. Experimental results on five classification benchmarks suggest the effectiveness of our proposed method. Further statistical and visualized analysis also explicitly shows the clustering results and provides interpretability for the classification results. For future study, we can try out other encoding basement for capturing the words' semantics and we are interested in considering phrases instead of individual words as the basic elements for clustering. The idea of clustering representation is worth trying on other NLP tasks.

\bibliographystyle{named}
\bibliography{ijcai19}

\end{document}